\documentclass{article}
\usepackage[a4paper,margin=1in]{geometry}
\usepackage{authblk}
\usepackage{hyperref}
\usepackage{orcidlink}
\usepackage{amsmath, amssymb, amsfonts, graphicx, subcaption}

\usepackage[linesnumbered,ruled,vlined]{algorithm2e}
\usepackage{wrapfig}
\usepackage{url}
\usepackage{accents}

\newcommand{\ubar}[1]{\underaccent{\bar}{#1}}
\newcommand{\customsubfigure}[3]{%
    \begin{subfigure}[b]{0.495\textwidth}
        \includegraphics[width=\textwidth]{#2}
    \end{subfigure}
    \begin{subfigure}[b]{0.495\textwidth}
        \includegraphics[width=\textwidth]{#3}
    \end{subfigure}
    \vspace{-1em}
    \caption*{#1}
}

\title{Data-Parallel Neural Network Training via Nonlinearly Preconditioned Trust-Region Method}

\author[1,2]{Samuel A. Cruz Alegr\'ia \orcidlink{0009-0004-5144-256X}}
\author[1]{Ken Trotti \orcidlink{0000-0002-5496-9445}}
\author[3,1]{Alena Kopani\v{c}\'akov\'a \orcidlink{0000-0001-8388-5518}}
\author[1,2]{Rolf Krause \orcidlink{0000-0001-5408-5271}}

\affil[1]{Universit\`a della Svizzera Italiana, Via la Santa 1, 6962, Viganello, Switzerland}
\affil[2]{UniDistance Suisse, Schinerstrasse 18, 3900, Brig, Switzerland}
\affil[3]{Brown University, 170 Hope Street, Providence, RI 02906, USA}

\date{}

\begin{document}
\maketitle

\begin{abstract}
Parallel training methods are increasingly relevant in machine learning (ML) due to the continuing growth in model and dataset sizes. We propose a variant of the Additively Preconditioned Trust-Region Strategy (APTS) for training deep neural networks (DNNs). The proposed APTS method utilizes a data-parallel approach to construct a nonlinear preconditioner employed in the nonlinear optimization strategy. In contrast to the common employment of Stochastic Gradient Descent (SGD) and Adaptive Moment Estimation (Adam), which are both variants of gradient descent (GD) algorithms, the APTS method implicitly adjusts the step sizes in each iteration, thereby removing the need for costly hyperparameter tuning. We demonstrate the performance of the proposed APTS variant using the MNIST and CIFAR-10 datasets. The results obtained indicate that the APTS variant proposed here achieves comparable validation accuracy to SGD and Adam, all while allowing for parallel training and obviating the need for expensive hyperparameter tuning.
\end{abstract}

\section{Introduction}
We consider the \textit{supervised learning problem}:
\begin{equation}\label{eq:min_problem}
\min_{\theta \in \mathbb{R}^n} f(\theta; \mathcal{D}) \;:=\; \min_{\theta \in \mathbb{R}^n} \frac{1}{p} \sum_{i=1}^p \ell(\mathcal{N}(x_i; \theta), c_i),
\end{equation}
where $f \colon \mathbb{R}^n \longrightarrow \mathbb{R}$ is the loss function, $\mathcal{D} \subset \mathbb{R}^{d_i} \times \mathbb{R}^{d_o}$ is a given dataset, $(x_i, c_i) \in \mathcal{D}$, and $\mathcal{N} \colon \mathbb{R}^{d_i} \times \mathbb{R}^n \longrightarrow \mathbb{R}^{d_o}$ is a model with parameters $\theta \in \mathbb{R}^n$. 
The metric $\ell \colon \mathbb{R}^{d_o} \times \mathbb{R}^{d_o} \longrightarrow \mathbb{R}$ measures the discrepancy between the \textit{prediction} $\mathcal{N}(x_i; \theta)$ and the \textit{target} $c_i$, where $c_i$ is the ``ground truth". 
In ML, $\mathcal{N}$ is typically a neural network (NN), and \textit{training} is the process of finding parameters $\theta$ which approximately minimize $f$. 
Thus, the dataset $\mathcal{D}$ is often termed as the \textit{training set} and $f(\theta; \mathcal{D})$ as the \textit{training loss}.

In the ML field, two popular training algorithms are SGD and Adam \cite{kingmaAdam2014}. Although SGD provides efficiency due to its computationally cheap gradient estimates, identifying appropriate hyperparameters is computationally costly and time-consuming (\textit{hyperparameter tuning}). In particular, the step size, or learning rate, depends heavily on the respective dataset and NN architecture. Moreover, SGD provides global convergence guarantees only under very restrictive conditions, e.g. a diminishing learning rate \cite{bottouNocedal2018}. Adam overcomes this constraint to some extent through its estimates of the first and second moments, but it does not provide convergence guarantees for nonconvex problems \cite{kingmaAdam2014}, and may even face issues when confronted with convex problems \cite{reddiEtAl2018OpenReview}. 
In the field of optimization, Trust-Region (TR) \cite{connGouldToint2000} is a widely-used method whose design guarantees global convergence for both convex and nonconvex problems, all without expensive hyperparameter tuning.
For this reason, TR has also begun to gain a foothold in ML; see, for example,~\cite{kopanickovaKrause2022, gratton2023multilevel}. Recently, parallel training approaches have gained significance as datasets and DNN architectures continue to grow in size and complexity; see, for example, the survey \cite{nicholsEtAl2021} and the references therein.

In general, we can divide parallel training methods into two main classes: model- and data-parallel approaches.
Model-parallel approaches distribute the NN across multiple processors, either by independently training distinct layers or by dividing the model into segments composed of various layer subsets. 
Indeed, several model-parallel frameworks are available, e.g. \cite{ rasleyEtAl2020}. 

Data-parallel approaches instead distribute the training set $\mathcal{D}$, or the current minibatch, across $N$ processing units, e.g. CPUs and GPUs. Each unit then processes a subset of $\mathcal{D}$ and updates a shared model or a copy of a ``main NN". 
Popular classes of data-parallel methods include federated learning methods and ensemble methods.
On the one hand, federated learning is an approach in which a model is trained across multiple decentralized processing units holding local minibatches without exchanging them \cite{mcmahanEtAl2017}. On the other hand, ensemble methods combine the locally obtained estimates of loss and its derivatives. These local estimates are typically synchronized and averaged to obtain the final estimate, but other techniques also exist~\cite{dietterich2000}. 

Using such principles, parallel variants of SGD have been devised {\cite{zinkevichEtAl2010, zhangEtAl2015}. 
In \cite{zinkevichEtAl2010}, the authors introduce Parallel SGD, which enables (synchronous) parallel NN training through a data-parallel approach. 
However, convergence guarantees are still constrained by a fixed and ``low enough" learning rate. 
In \cite{zhangEtAl2015}, the authors introduce Elastic Averaging SGD, which allows for parallel synchronous and asynchronous NN training via a data-parallel approach. 
In this case, the ``main NN" parameters are updated via a moving average of the local NN parameters. 

The convergence of this approach is only proven for the quadratic and strongly convex case, which is not typically the nature of \eqref{eq:min_problem}. 

In this work, we develop a data-parallel Additively Preconditioned Trust-Region Strategy (APTS)}, which is globally convergent, inherently parallelizable, and does not require hyperparameter tuning. Compared to standard data-parallel approaches, the devised APTS method trains local networks in parallel. This, in turn, reduces the communication cost associated with the synchronization step compared to standard data-parallel approaches. We emphasize that our method is inspired by additive Schwarz domain decomposition (DD) methods and TR. DD methods were initially developed to solve Partial Differential Equations (PDEs) numerically \cite{chanEtAl1994, toselliWidlund2004, gander2006, erhelEtAl2014, mathew2008}.
Notable examples of additive DD methods include: Additive Schwarz Preconditioned Inexact Newton algorithm (ASPIN) \cite{caiKeyes2002}, Globalized ASPIN (GASPIN) \cite{grossKrause2021}, Restricted Additive Schwarz Preconditioned Exact Newton (RASPEN) \cite{doleanEtAl2016}, and APTS \cite{grossKrause2009}. 
Most of these methods were designed to solve non-linear systems of equations, while APTS and GASPIN were developed to solve nonconvex minimization problems. Both approaches, however, are closely related, as the necessary first-order conditions for a minimization problem constitute a (non-)linear system of equations. 

DD methods have recently been applied to ML.
In \cite{guEtAl2022, guEtAl2023}, the authors explore subnetwork transfer learning via the decomposition and composition of deep convolutional NNs (CNNs). 
In \cite{kopanickovaEtAl2023}, the authors employ a non-linear and layer-wise right preconditioner for the limited-memory Broyden–Fletcher–Goldfarb–Shanno (L-BFGS) algorithm in the context of physics-informed NNs (PINNs). 
In \cite{klawonnEtAl2023}, the authors explore a DD-inspired model-parallel training method for CNNs designed for image recognition tasks. 
Finally, in \cite{trottiEtAl2023}, we applied APTS to the NN parameter space in a layer-parallel fashion similar to that used in \cite{kopanickovaEtAl2023}. 
This transition from solving PDEs to enhancing NN training demonstrates the versatility and applicability of DD methods.\\

\section{Data-Parallel Additively Preconditioned Trust Region}
\subsection{Trust-Region Method}
\label{subsec:tr_method}
TR methods \cite{connGouldToint2000} are iterative methods that provide global convergence to a local minimizer by adapting their step size depending on the quality of a model $m \colon \mathbb{R}^n \longrightarrow \mathbb{R}$. 
At each iteration $k$, the model $m$ is constructed using the first-order, and possibly the second-order, Taylor expansion around the current iterate, i.e.~the parameters $\theta^k$. 
The search direction $s^k$ is obtained by solving the following constrained quadratic minimization problem:
\begin{equation}
\label{eq:tr_subproblem}
\min_{\{s^k \in \mathbb{R}^n~|~\| s^k \| \leq \Delta^k\}} m(s^k) :=
\nabla f(\theta^k; \mathcal{D})^T (\theta^k+s^k) + \frac{1}{2} (\theta^k + s^k)^T B^k (\theta^k+s^k),
\end{equation}
where $\Delta^k \in \mathbb{R}$ is an upper bound for the step length, and it is known as the \textit{trust-region radius}. 
The operator $B \in \mathbb{R}^{n \times n}$ is either a Hessian of~$f$ or its approximation. After~$s^k$ is obtained, its quality is assessed by the TR algorithm, which determines its acceptance and provides guidance for adjusting the TR radius accordingly.

In ML, especially for large NNs, computations with the exact Hessian become intractable. 
Hence, we resort to its L-SR1 approximation~\cite{ erwayMarcia2015}, which, in contrast to the L-BFGS approximation~\cite{nocedalWright1999}, allows for indefinite Hessian approximations.
Moreover, to solve the TR subproblem \eqref{eq:tr_subproblem}, we apply the Orthonormal Basis SR1 (OBS) method from \cite{brust2017, erway2020}. 
The L-SR1 method keeps track of the $m$ most recent secant pairs $\{ (s^i, y^i) \}_{i=k-m}^k$, where each $s^k$ is obtained by solving~\eqref{eq:tr_subproblem} and 
$y^k := \nabla f(\theta^{k} +s^k ; \mathcal{D}) - \nabla f(\theta^k; \mathcal{D})$. 
Traditionally, in the minibatch regime, evaluating $y^k$ using the full dataset $\mathcal{D}$ becomes computationally expensive and, therefore, $y^k$ is evaluated only using a small portion of the data constituting an overlap, which is imposed between two consecutive minibatches.

\subsection{Data-Parallel Additively Preconditioned Trust Region}
\label{subsec:apts_d}
APTS \cite{grossKrause2009} is a right-preconditioned additive DD method which provides global convergence guarantees due to its use of a TR globalization strategy. 

To formalize this method, we begin by noting that the training set $\mathcal{D}$ is split into $N$ minibatches $\{\mathcal{D}_i\}_{i=1}^N$. Each minibatch is used to train, in parallel, one of $N$ copies of the neural network $\mathcal{N}$ with parameters $\theta$. Thus, we denote the $i$-th network copy by $\mathcal{N}_i(\theta, \mathcal{D}_i)$. 
Overlap between minibatches is allowed, i.e. we permit for either $\mathcal{D}_i \cap \mathcal{D}_j = \emptyset$ or $\mathcal{D}_i \cap \mathcal{D}_j \neq \emptyset$, for $i \neq j$.
Following our nomenclature, we will refer to objects related to $\mathcal{D}$ and $\mathcal{D}_i$ as ``global" and ``local" objects, respectively. 

Let index $i$ denote the minibatch/network (which we here refer to as a subdomain), while indices $j$ and $k$ are used to indicate the local TR iteration and the global APTS iteration, respectively.
We define the local training loss as follows
\begin{equation}
    \label{eq:local_training_loss}
    f_i(\theta^k_{i,j}, \mathcal{D}_i) := f(\theta^k_{i,j}, \mathcal{D}_i) + \langle r^k_i , \theta^k_{i,j} - \theta^k_{i,0}\rangle,
\end{equation}
where the first-order consistency term $r_i^k$ is defined as 
\begin{equation}
 {r_i^k := \nabla f (\theta^k, \mathcal{D}) - \nabla f(\theta^k_{i,0}, \mathcal{D}_i)}.   
\end{equation}
Thus, the local training loss $f_i$ is defined so that the local gradient equals the global gradient in the first local TR iteration. 
This, in turn, leads to variance reduction of the minibatch gradient; see \cite{braglia2020Multilevel} for details. 

Algorithm \ref{alg:APTS_D} provides an overview of our proposed variant of APTS developed for data-parallel training of NNs. 
We have also devised a stochastic variant termed SAPTS. 
In the deterministic case (APTS), the dataset $D$ (c.f. Algorithm \ref{alg:APTS_D}, line \ref{alg:stoch_or_not}) corresponds to the full dataset, and the subsets $\mathcal{D}_i$ can be seen as minibatches. 
In the stochastic case (SAPTS), the set $D$ corresponds to a minibatch, while $\mathcal{D}_i$ are subsets of the minibatch, which we call \textit{microbatches}. 
In case they are overlapping, we apply an overlap between \textit{all} microbatches of a minibatch. 
We note that the existing global convergence analysis of APTS does not directly extend to SAPTS, and adjustments will be made in the future. 

On line \ref{alg:initialization}, we initialize the TR radii $\Delta^k_{i,0}$ and maximum TR radii $\bar{\Delta}^{k}_{i,0}$ such that the global trial step $\tilde{s}^k := \sum_{i=1}^N s_i^k$ is bounded by the global TR radius $\Delta^k$, i.e.,
\begin{equation}
\|\tilde{s}^k\|_2 = \|\sum_{i=1}^N s_i^k\|_2 \leq N \|\frac{\Delta^k}{N}\|_2 \leq \Delta^k,
\end{equation}
where $s_i^k\,, i = 1,\ldots, N$, are the local steps.

The preconditioning step on line \ref{alg:evaluation} consists of training the NN copies using the local TR method, which is executed until the iteration count reaches the maximum amount $\nu$ or until the local step $s_i^k$ reaches the boundary of the local trust region. 
After the preconditioning step, on line \ref{alg:evaluation}, the quality of the global and local updates is assessed through the ratio:
\begin{equation}
    \rho^k = \frac{f(\theta^k, \mathcal{D}) - f(\theta^k + \tilde{s}^k, \mathcal{D})}{\sum_{i=1}^{N} (f_i(\theta^k_{i,0}, \mathcal{D}_i) - f_i(\theta^k_{i,0} + s^k_i, \mathcal{D}_i))},
    \label{eq:decrease_ratio}
\end{equation}
where $f$ and $f_i$ are the global and local training losses defined in \eqref{eq:min_problem} and \eqref{eq:local_training_loss}, respectively. 
This ratio compares the average decrease of the local loss functions observed across all local networks with respect to the decrease in the global loss function after considering the accumulated local corrections.
If $\rho^k > \eta$, this indicates a good agreement between the global and local NNs. 
This, in turn, allows for an increase of the TR radius and the adoption of the new parameters for the global TR iteration on line \ref{alg:global_step}. 
Otherwise, the TR radius is left unchanged or decreased, and the global parameters are also left unchanged. 
In this fashion, on line \ref{alg:temp_update}, the global parameters and TR radius are updated as follows:
\begin{equation}
    \tilde{\theta}^k = 
    \begin{cases}
    \theta^k + \tilde{s}^k, & \text{ if } \rho^k \geq \eta_1,\\
    \theta^k, & \text{ otherwise}
    \end{cases}
    ,
    \hspace{1cm}
    \tilde{\Delta}^k = 
    \begin{cases}
    \min(\beta \Delta^k, \bar{\Delta}), & \text{ if } \rho^k > \eta_2,\\
    \max(\alpha \Delta^k, \ubar{\Delta}), & \text{ if } \rho^k < \eta_1,\\
    \Delta^k, & \text{ otherwise},
    \end{cases}
    \label{eq:temporary_update}
\end{equation}
where $0 < \alpha < 1$ is the decrease factor, $\beta > 1$ is the increase factor, $\ubar{\Delta}, \bar{\Delta}$ are the minimum and maximum TR radii, respectively, and $0 \leq \eta_1 < \eta_2 < 1$ establish the criteria for updating. 
Subsequently, on line \ref{alg:global_step}, a single TR iteration is executed on the global network $\mathcal{N}(\tilde{\theta}^k, \mathcal{D})$, with $\tilde{\Delta}^k$ as the global TR radius. 
This yields the final global parameters $\theta^{k+1}$ and TR radius $\Delta^{k+1}$.
This whole procedure is repeated until a stopping criterion has been met.

\begin{algorithm}
\caption{APTS in data}\label{alg:APTS_D}

\DontPrintSemicolon
\KwIn{$f \colon \mathbb{R}^n \longrightarrow \mathbb{R}$, $f_i \colon \mathbb{R}^n \longrightarrow \mathbb{R}$, $\mathcal{D} \subset \mathbb{R}^{d_i} \times \mathbb{R}^{d_o}$, $\mathcal{N}(\theta^0, \mathcal{D}) \colon \mathbb{R}^{d_i} \times \mathbb{R}^n \longrightarrow \mathbb{R}^{d_o}$, $\nu, n_{\text{epochs}} \in \mathbb{N}$, $\Delta^0, \ubar{\Delta}, \bar{\Delta} \in \mathbb{R}_{\geq 0}$, $0 < \eta_1 < \eta_2 < 1$, $0 < \alpha < 1 < \beta$}
\For{$\mathrm{epoch}=1, \ldots, n_{\mathrm{epochs}}$}{
            $k \gets 0$\;
            \For{$D \in [\mathcal{D}]$ $(\mathrm{APTS})$ or $D\in[D_1, D_2, \ldots]$ $(\mathrm{SAPTS})$}{ \label{alg:stoch_or_not}
        \For{$i \in [1, \ldots, N]$ $\mathrm{in~parallel}$}{
            $\theta_{i,0}^k\gets \theta^k$,\quad $\Delta_{i,0}^k\gets \Delta^k/N$, 
            \quad$\ubar{\Delta}^{k}_{i,0} = \ubar{\Delta}/N$,
            \quad$\bar{\Delta}^{k}_{i,0} = \bar{\Delta}/N$ \label{alg:initialization}\; 
            Using $f_i$ as defined in \eqref{eq:local_training_loss}, apply TR to $\mathcal{N}_i(\theta_{i,0}^k, D)$, yielding $s_i^k$ \label{alg:preconditioning}\;
        }
        Evaluate $\rho^k$ as in \eqref{eq:decrease_ratio} \label{alg:evaluation}\;
        Compute $\tilde{\theta}^k$ and $\tilde{\Delta}^k$ as in \eqref{eq:temporary_update} \label{alg:temp_update}\;
        Using $f$ as defined in \eqref{eq:min_problem}, apply TR to $\mathcal{N}(\tilde{\theta}^k, D)$, yielding $\theta^{k+1}$ and $\Delta^{k+1}$ \label{alg:global_step}\;
        $k \gets k + 1$\;
    }
}
\end{algorithm}

\section{Numerical Results}
Our code \cite{cruzas_ML_APTS} is written in Python and leverages the PyTorch framework \cite{paszke2019}. 
All APTS tests were run on the Piz Daint supercomputer at the Swiss National Supercomputing Centre (CSCS), on XC50 compute nodes---Intel® Xeon® E5-2690 v3 @ 2.60GHz (12 cores, 64GB RAM) and NVIDIA® Tesla® P100 16GB. 
We utilized PyTorch's \texttt{DistributedDataParallel} library, which streamlined the management of the network copies discussed in Subsection \ref{subsec:apts_d}. 
We compare APTS with the sequential Adam optimizer, as SGD and TR performed less optimally in terms of validation accuracy. 
The results for Adam are the best results achieved based on the average validation accuracy obtained after 10 trials across 13 learning rates. 
Adam's results may surpass APTS in some cases; however, this advantage requires expensive fine-tuning for each problem and network architecture, unlike APTS, which uses constant parameters without such tuning. 
We tested our APTS implementation on the MNIST and CIFAR-10 datasets, using either the full dataset (60'000 and 50'000 training samples, respectively) or minibatches of size 10'000. 
Additionally, we used a fully connected NN with 535'818 parameters and 1'841'162 parameters in 3 layers (6 PyTorch tensors) for MNIST and CIFAR-10, respectively. 
In the minibatch experiments, we introduced an overlap of 5\% between all minibatches and the microbatches belonging to each minibatch. 
Finally, we used a maximum of five local TR iterations.

\begin{figure}
    \centering
    \customsubfigure{}{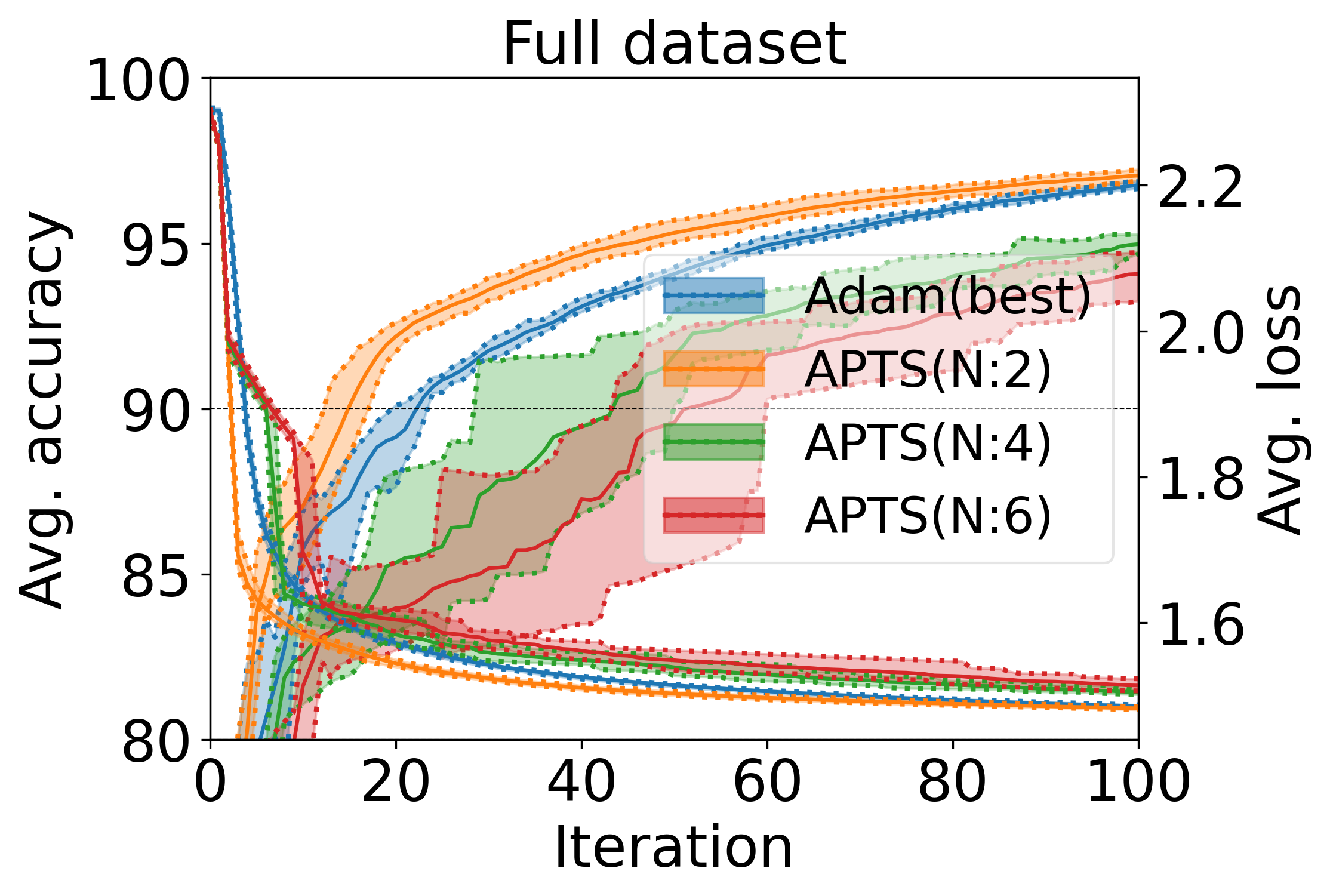}{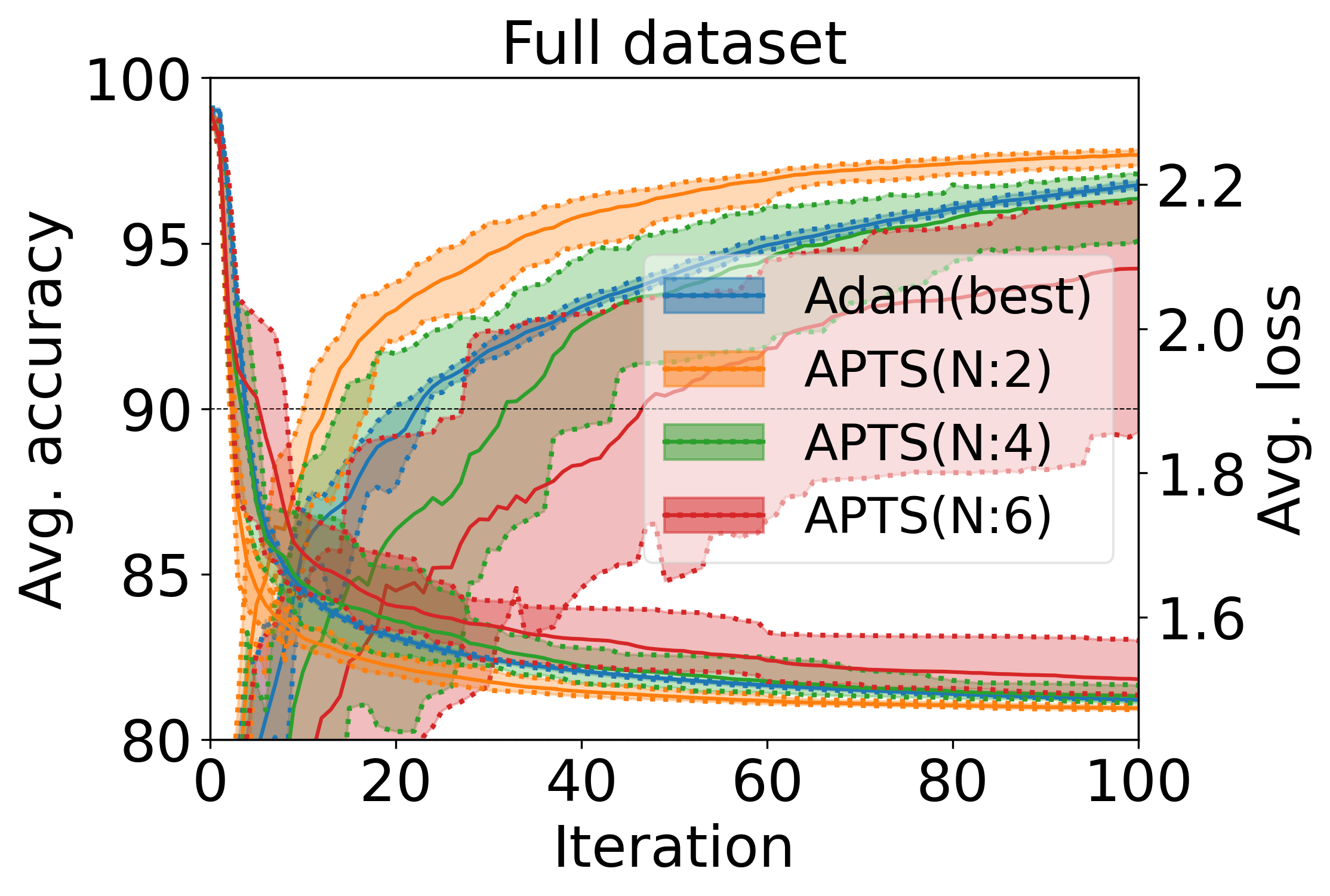}
    \vspace{-2em}
    \customsubfigure{}{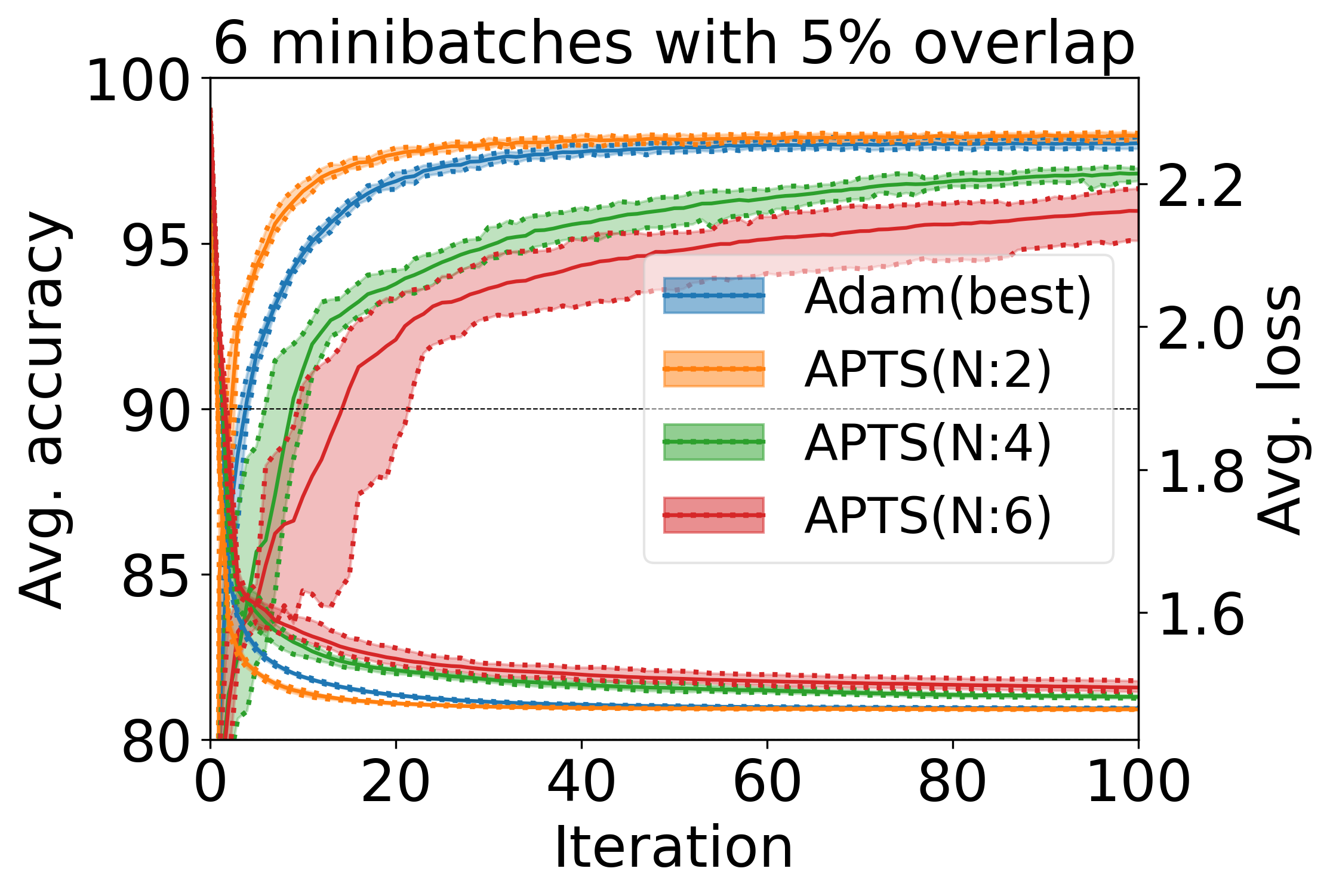}{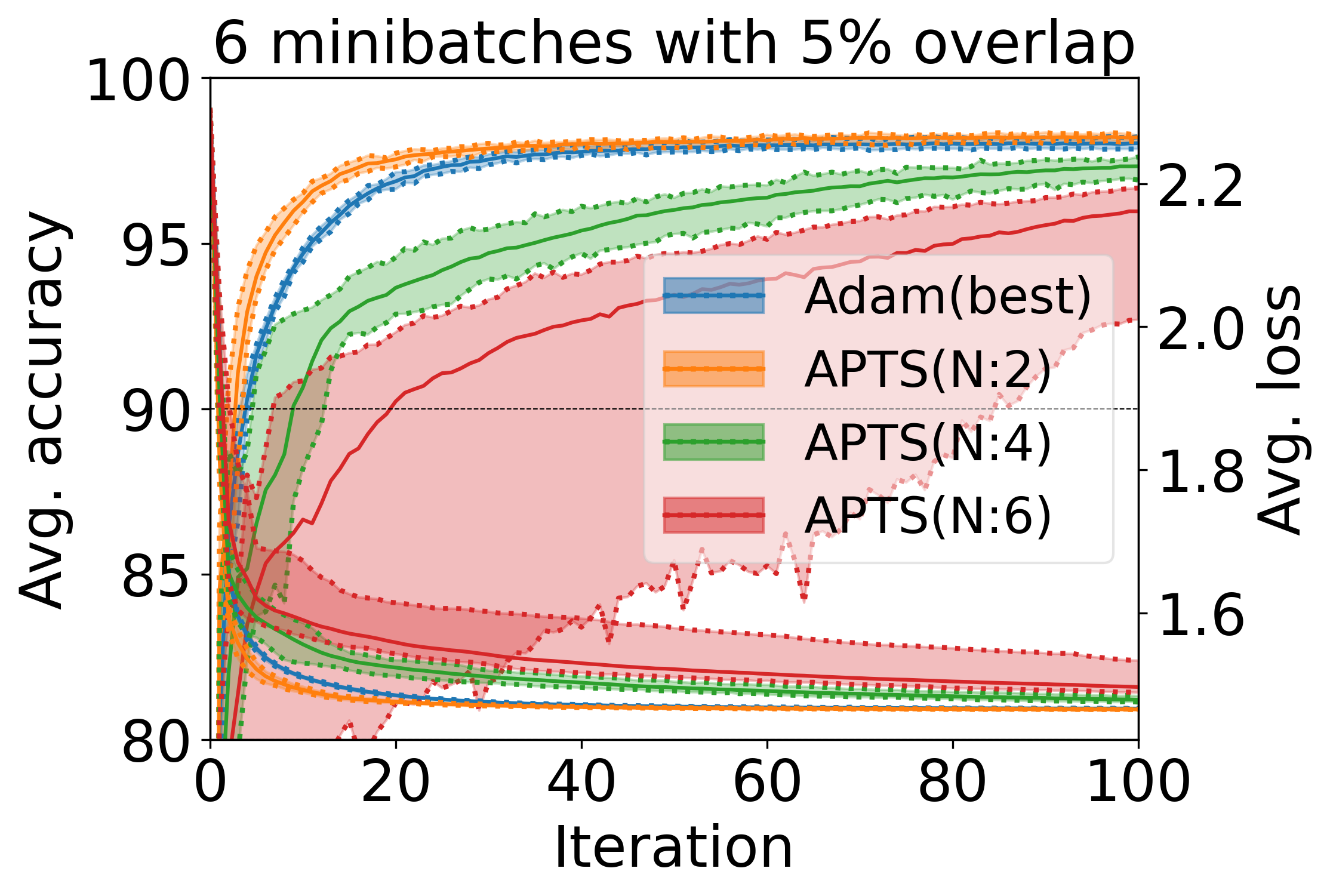}
    \vspace{-2em}
    \caption{MNIST. Left: first-order model, right: second-order model.}
    \label{fig:mnist_apts_d}
\end{figure}

\noindent In Figure \ref{fig:mnist_apts_d}, APTS demonstrates competitiveness with the fine-tuned Adam algorithm for the MNIST example. 
Notably, by minimizing loss as the primary metric, APTS achieves a loss value comparable to Adam's. 
However, it's evident that the accuracy diminishes slightly as the number of subdomains increases. 
This is expected since the locally computed gradients are less accurate approximations of the full gradient as the data volume decreases. 
Furthermore, there's a noticeable reduction in variance across all experiments when only first-order information is used in the TR subproblem \eqref{eq:tr_subproblem}, compared to when both first- and second-order information is utilized. 

This can most likely be attributed to the instabilities in Hessian approximations, which are caused by utilizing small overlaps between all minibatches and microbatches.
We plan to investigate this drawback in the future.

\begin{figure}
    \customsubfigure{}{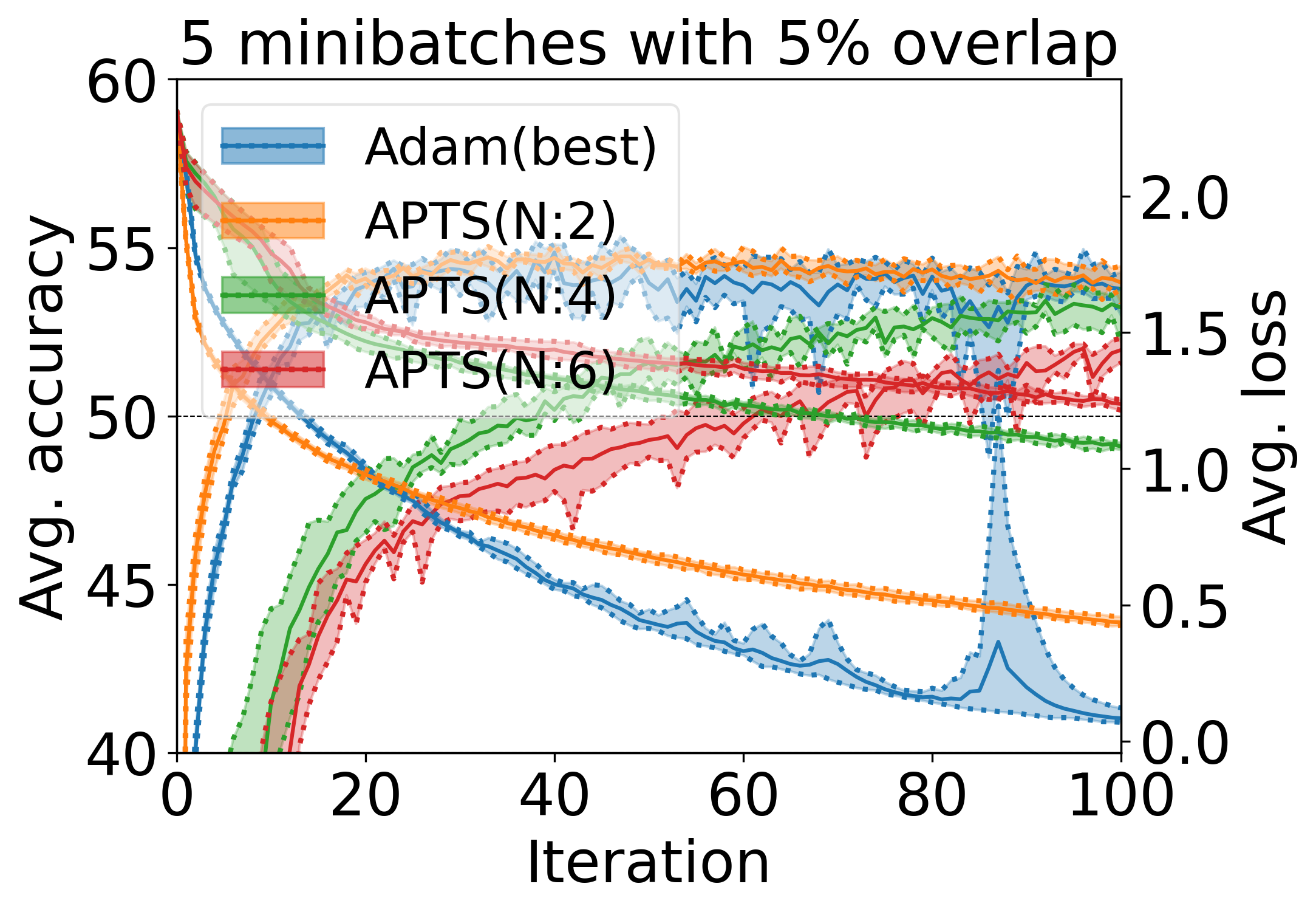}{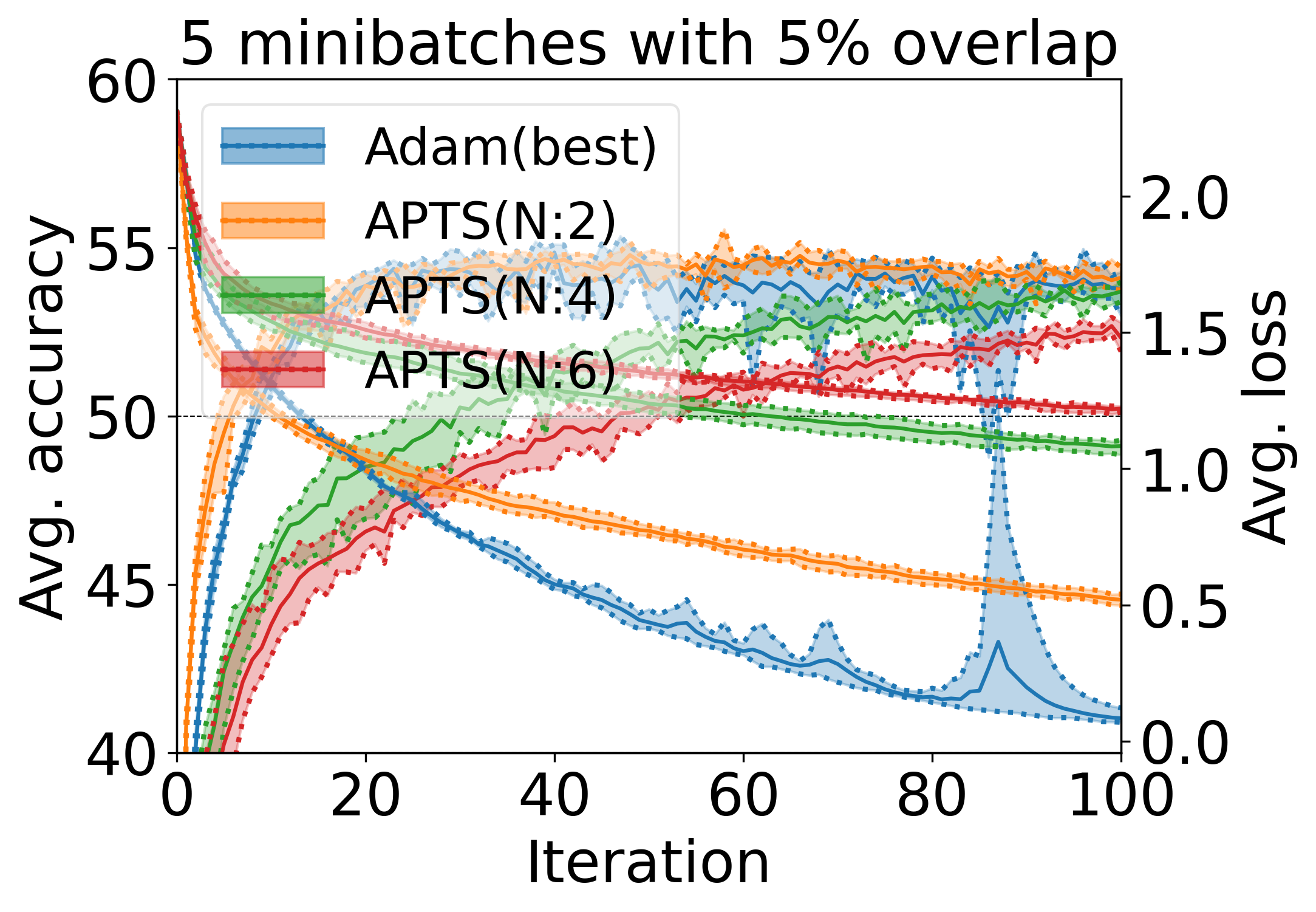}
    \vspace{-2em}
    \caption{CIFAR-10. Left: first-order model, right: second-order model.}
  \label{fig:cifar10_apts_d}
\end{figure}

In Figure \ref{fig:cifar10_apts_d}, APTS showcases performance comparable to Adam's achieved accuracy for the CIFAR-10 example. 
We notice that Adam appears to plateau, possibly due to overfitting, as evidenced by its loss nearing zero. 
Moreover, Adam demonstrates an interesting accuracy artifact: a deep and shallow peak close to the end of training. 
In contrast, APTS continues to improve steadily and outperforms Adam in terms of accuracy. This suggests that while Adam may initially be faster in achieving a certain accuracy, APTS may be able to provide a more steady and robust improvement in NN training. Regarding parallel performance, more subdomains in APTS led to a decrease in time-to-solution. Each variant performed at most five local TR iterations, after which the processes shared information among themselves, so the communication overheard depends only on the connection speed between the different nodes. This contrasts with parallel versions of SGD and Adam, in which a synchronization step is performed after each loss/gradient evaluation. Thus, the proposed APTS method is expected to scale better than naive data-parallel approaches due to reduced communication overhead. We plan to perform empirical studies regarding the scalability of the proposed method in the near future.

\section{Conclusion}
In this work, we introduced a data-parallel variant of the APTS method that eliminates the need for hyperparameter tuning. 
Our experiments demonstrate that this method achieves validation accuracy comparable to the Adam optimizer, particularly when training sets are split into two minibatches. 
At the same time, the devised APTS algorithm provides an opportunity for parallel data training at a reduced communication cost compared to standard data-parallel approaches. The numerical results show that APTS is scalable only for a smaller number of subdomains. However, it may be combined with a decomposition in the parameter space \cite{trottiEtAl2023}. 

\section*{Acknowledgement}
This work was supported by the Swiss Platform for Advanced Scientific Computing (PASC) project ExaTrain (funding periods 2017-2021 and 2021-2024) and by the Swiss National Science Foundation through the projects ``ML$^2$ -- Multilevel and Domain Decomposition Methods for Machine Learning'' (197041) and ``Multilevel training of DeepONets -- multiscale and multiphysics applications'' (206745).


\begin{thebibliography}{40.}%

\bibitem{cruzas_ML_APTS} Cruz Alegría, S.A., Trotti, K.: ML\_APTS. GitHub repository. Retrieved from \url{https://github.com/cruzas/ML_APTS} (2023).

\bibitem{trottiEtAl2023} Trotti, K., Cruz Alegría, S.A., Kopaničáková, A., Krause, R: Parallel Trust-Region Approaches in Neural Network Training: Beyond Traditional Methods. arXiv preprint 2312.13677 (2023).

\bibitem{chanEtAl1994} Chan, T.F., Zou, J.: Additive Schwarz domain decomposition methods for elliptic problems on unstructured meshes. Numerical Algorithms, \textbf{8}, 329–346 (1994).

\bibitem{nocedalWright1999} Nocedal, J., Wright, S.J. (eds.): Numerical Optimization. Springer, New York (1999).

\bibitem{connGouldToint2000} Conn, A.R., Gould, N.I., Toint, P.L.: Trust Region Methods. Society for Industrial and Applied Mathematics (2000).

\bibitem{dietterich2000} Dietterich, T.G.: Ensemble methods in machine learning. In: International Workshop on Multiple Classifier Systems, pp. 1–15. Springer, Berlin Heidelberg (2000).

\bibitem{caiKeyes2002} Cai, X.C., Keyes, D.E.: Nonlinearly preconditioned inexact newton algorithms. SIAM Journal on Scientific Computing, \textbf{24}, 183–200 (2002).

\bibitem{toselliWidlund2004} Toselli, A., Widlund, O.: Domain Decomposition Methods - Algorithms and Theory, vol. 34. Springer (2004). 

\bibitem{gander2006} Gander, M.J.: Optimized Schwarz methods. SIAM Journal on Numerical Analysis, \textbf{44}, 699–731 (2006).

\bibitem{mathew2008} Mathew, T.P.: Domain Decomposition Methods for the Numerical Solution of Partial Differential Equations. Springer (2008). 

\bibitem{grossKrause2009} Groß, C.: A Unifying Theory for Nonlinear Additively and Multiplicatively Preconditioned Globalization Strategies Convergence Results and Examples From the Field of Nonlinear Elastostatics and Elastodynamics. Ph.D. thesis, Bonn International Graduate School, University of Bonn (2009).

\bibitem{zinkevichEtAl2010} Zinkevich, M., Weimer, M., Li, L., Smola, A.: Parallelized stochastic gradient descent. In: Advances in Neural Information Processing Systems, vol. 23 (2010). 

\bibitem{erhelEtAl2014} Erhel, J., Gander, M.J., Halpern, L., Pichot, G., Sassi, T., Widlund, O. (eds.): Domain Decomposition Methods in Science and Engineering XXI. Springer International Publishing (2014).

\bibitem{kingmaAdam2014} Kingma, D.P., Ba, J.: Adam: A method for stochastic optimization. arXiv preprint 1412.6980 (2014).

\bibitem{erwayMarcia2015} Erway, J.B., Marcia, R.F.: On efficiently computing the eigenvalues of limited-memory quasi-Newton matrices. SIAM Journal on Matrix Analysis and Applications, \textbf{36}, 1338–1359 (2015).

\bibitem{zhangEtAl2015} Zhang, S., Choromanska, A.E., LeCun, Y.: Deep learning with elastic averaging SGD. In: Advances in Neural Information Processing Systems, vol. 28 (2015).

\bibitem{doleanEtAl2016} Dolean, V., Gander, M.J., Kheriji, W., Kwok, F., Masson, R.: Nonlinear Preconditioning: How to use a nonlinear Schwarz method to precondition Newton's method. SIAM Journal on Scientific Computing, \textbf{38}, A3357–A3380 (2016). 

\bibitem{brust2017} Brust, J., Erway, J.B., Marcia, R.F.: On solving L-SR1 trust-region subproblems. Computational Optimization and Applications, \textbf{66}, 245–266 (2017).

\bibitem{mcmahanEtAl2017} McMahan, B., Moore, E., Ramage, D., Hampson, S., y Arcas, B.A.: Communication efficient learning of deep networks from decentralized data. In Artificial Intelligence and Statistics, pp. 1273–1282 (2017). 

\bibitem{bottouNocedal2018} Bottou, L., Curtis, F.E., Nocedal, J.: Optimization methods for large-scale machine learning. SIAM Review, \textbf{60}, 223–311 (2018). 

\bibitem{reddiEtAl2018OpenReview} Reddi, S.J., Kale, S., Kumar, S.: On the convergence of Adam and beyond. arXiv preprint 1904.09237 (2019).

\bibitem{paszke2019} Paszke, Adam, et al.: Pytorch: An imperative style, high-performance deep learning library. In: Advances in Neural Information Processing Systems, vol. 32 (2019).

\bibitem{braglia2020Multilevel} Braglia, V., Kopaničáková, A., Krause, R.: A multilevel approach to training. arXiv preprint 2006.15602 (2020).

\bibitem{erway2020} Erway, J.B., Griffin, J., Marcia, R.F., Omheni, R.: Trust-region algorithms for training responses: machine learning methods using indefinite Hessian approximations. In: Optimization Methods and Software, vol. 35, pp. 460–487 (2020).

\bibitem{rasleyEtAl2020} Rasley, J., Rajbhandari, S., Ruwase, O., He, Y.: DeepSpeed: System Optimizations Enable Training Deep Learning Models with Over 100 Billion Parameters. In: Proceedings of the 26th ACM SIGKDD International Conference on Knowledge Discovery \& Data Mining, pp. 3505–3506 (2020).

\bibitem{grossKrause2021} Groß, C., Krause, R.: On the Globalization of ASPIN Employing Trust-Region Control Strategies--Convergence Analysis and Numerical Examples. arXiv preprint 2104.05672 (2021).

\bibitem{nicholsEtAl2021} Nichols, D., Singh, S., Lin, S.H., Bhatele, A.: A Survey and Empirical Evaluation of Parallel Deep Learning Frameworks. arXiv preprint 2111.04949 (2021).

\bibitem{kopanickovaKrause2022} Kopaničáková, A., Krause, R.: Globally Convergent Multilevel Training of Deep Residual Networks. SIAM Journal on Scientific Computing, \textbf{0}, S254–S280 (2022). 

\bibitem{guEtAl2022} Gu, L., Zhang, W., Liu, J., Cai, X.C.: Decomposition and composition of deep convolutional neural networks and training acceleration via sub-network transfer learning. ETNA — Electronic Transactions on Numerical Analysis (2022).

\bibitem{gratton2023multilevel} Gratton, S., Kopaničáková, A., Toint, P.L.: Multilevel Objective-Function-Free Optimization with an Application to Neural Networks Training. SIAM Journal on Optimization, \textbf{33}, 2772–2800 (2023). 

\bibitem{guEtAl2023} Gu, L., Zhang, W., Liu, J., Cai, X.C.: Decomposition and Preconditioning of Deep Convolutional Neural Networks for Training Acceleration. In: Domain Decomposition Methods in Science and Engineering XXVI, pp. 153–160. Springer International Publishing (2023). 

\bibitem{klawonnEtAl2023} Klawonn, A., Lanser, M., Weber, J.: A Domain Decomposition-Based CNN-DNN Architecture for Model Parallel Training Applied to Image Recognition Problems. arXiv preprint 2302.06564 (2023).

\bibitem{kopanickovaEtAl2023}
Kopaničáková, A., Kothari, H., Karniadakis, G.E., Krause, R.: Enhancing Training of Physics-Informed Neural Networks Using Domain Decomposition–Based Preconditioning Strategies. SIAM Journal on Scientific Computing, pp. S46–S67 (2024).

\end{thebibliography}
\end{document}